\documentclass[10pt,conference]{IEEEtran}
\IEEEoverridecommandlockouts
\usepackage{amsmath,amssymb,amsfonts}
\usepackage{algorithmic}
\usepackage{graphicx}
\usepackage{textcomp}
\usepackage{xcolor}
\usepackage{xspace}
\usepackage{enumitem}
\usepackage{pifont}
\usepackage{bbding}
\usepackage{algorithm}
\usepackage{multirow}
\usepackage{caption}
\usepackage{flushend}
\usepackage{bbding}
\usepackage{threeparttable}

\usepackage{setspace}
\renewcommand\baselinestretch{0.91}

\usepackage[square,sort,comma,numbers,compress]{natbib}

\begin{document}

\title{\Large\bf BSC: Block-based Stochastic Computing to Enable Accurate and Efficient TinyML}
\author{
\IEEEauthorblockN{
Yuhong Song$^{1}$ \quad
Edwin Hsing-Mean Sha$^{1, * \thanks{* Edwin Hsing-Mean Sha is the corresponding author (edwinsha@cs.ecnu.edu.cn)}}$ \quad
Qingfeng Zhuge$^{1}$ \quad
Rui Xu$^{1}$ \quad
Yongzhuo Zhang$^{1}$ \quad \\
Bingzhe Li$^{2}$ \quad 
Lei Yang$^{3}$ \quad 
}
\vspace{5px}
\IEEEauthorblockA{
\normalsize
$^{1}$ East China Normal University \quad
$^{2}$ Oklahoma State University \quad
$^{3}$ University of New Mexico
}
}
\maketitle

\begin{abstract}
Along with the progress of AI democratization, machine learning (ML) has been successfully applied to edge applications, such as smart phones and automated driving.
Nowadays, more applications require ML on tiny devices with extremely limited resources, like implantable cardioverter defibrillator (ICD), which is known as TinyML.
Unlike ML on the edge, TinyML with a limited energy supply has higher demands on low-power execution.
Stochastic computing (SC) using bitstreams for data representation is promising for TinyML since it can perform the fundamental ML operations using simple logical gates, instead of the complicated binary adder and multiplier.
However, SC commonly suffers from low accuracy for ML tasks due to low data precision and inaccuracy of arithmetic units.
Increasing the length of the bitstream in the existing works can mitigate the precision issue but incur higher latency.
In this work, we propose a novel SC architecture, namely Block-based Stochastic Computing (BSC).
BSC divides inputs into blocks, such that the latency can be reduced by exploiting high data parallelism.
Moreover, optimized arithmetic units and output revision (OUR) scheme are proposed to improve accuracy.
On top of it, a global optimization approach is devised to determine the number of blocks, which can make a better latency-power trade-off.
Experimental results show that BSC can outperform the existing designs in achieving over 10\% higher accuracy on ML tasks and over 6$\times$ power reduction.

\end{abstract}

\setlength{\textfloatsep}{3pt}
\setlength{\floatsep}{1pt}
\setlength{\dbltextfloatsep}{3pt}
\setlength{\belowdisplayskip}{2pt}
\setlength{\abovedisplayskip}{2pt}


\section{Introduction}



With the rapid development of artificial intelligence (AI), deep neural networks (DNNs) as typical models of machine learning (ML) have been widely used on edge applications.
Recently, there are growing demands on the implementation of DNNs to tiny devices~\cite{kumar2017resource}, such as surveillance cameras in smart housing and implantable cardioverter defibrillator in intelligent healthcare, which is known as TinyML.
However, the energy supply (e.g., battery) of tiny devices is extremely limited, impeding the implementation of DNNs. Although many compression technologies
\cite{song2021dancing, jiang2020standing, peng2021accelerating,jiang2020hardware,yang2020co,jiang2019accuracy,qi2021accelerating}
through compacting models can save energy to a certain extent, it falls into low accuracy when the compression rate becomes high.
For TinyML, stochastic computing (SC) \cite{gaines1969stochastic, li2016using,li2019neural,li2018quantized,li2017neural} stands out 
to save energy by significantly simplifying computing circuits.

SC utilizes the probability of `1's in the whole bitstream to represent data.
For example, bitstream 0100 indicates 1/4 in unipolar format, but -2/4 (i.e., 2$\times$1/4-1) in bipolar format.
Regarding the data representation in SC, basic arithmetic operations such as addition and multiplication can be implemented by one simple gate instead of the traditional binary implementations with tens of gates~\cite{liu2020survey}.
For example, we can use an AND gate for unipolar and an XNOR gate for bipolar to approximate multiplication, and use an OR gate to approximate addition.
In contrast to conventional binary circuits, an SC circuit can significantly simplify hardware complexity and save energy.

However, SC usually suffers from low accuracy on basic ML operations (e.g., multiply-accumulate (MAC) and general matrix multiplication (GEMM)) and long computing latency.
The accuracy is affected by the design of arithmetic units and data precision.
Specifically for arithmetic units, the accuracy issue is mainly from two aspects. One is the correlation among inputs, which causes that the results of logic operating cannot be consistent with probability computing. The other is the overflow problem, it is ubiquitous when the accurate results exceed -1 or 1. Besides, existing works~\cite{najafi2018deterministic, faraji2019energy} increases the length of bitstreams to improve data precision, even using exponential bits to obtain accurate results. 
But, this will undoubtedly introduce long computing latency.

To deal with the long latency problem, in this work, we first divide bitstreams into blocks and execute them in parallel to reduce latency.
In this way, arithmetic units inside every block are required to have high accuracy.
The correlation and overflow may cause accuracy loss in every block.
Meanwhile, block division also brings inter-block inaccuracy problem.
To alleviate these accuracy problems, we propose optimized intra-block arithmetic units which can provide high computing accuracy.
Moreover, we develop the inter-block output revision (OUR) scheme to address missing `1's or redundancy `1's problem among blocks.
Finally, a novel strategy is devised to determine the number of block to make a better trade-off between latency and power.

The main contributions of this paper are listed as follows.
\begin{itemize}
    \item We propose a novel architecture, namely BSC, where the inputs are divided into blocks and executed them in parallel using optimized intra-block arithmetic units.
    \item A new output revision (OUR) scheme is designed to solve the inter-block inaccuracy problem. Besides, we first propose a novel heuristic strategy to guide block division for better latency-power trade-off.
    \item Comprehensive experiments are conducted and the results show that our methods can achieve higher accuracy on ML tasks than existing methods with the reduced latency and pipeline stalls, as well as over 6$\times$ power reduction compared with binary circuits.
\end{itemize}

The paper is organized as follows: In Section II, we present the background and motivations of this work. Section III presents our design.
Evaluation results are reported in Section IV. Section V concludes this paper.

\section{Background and Motivation}
\subsection{Related Work}
In order to obtain accurate results, long bitstreams are used to improve accuracy. Deterministic SC \cite{qian2010architecture} utilizes exponential bits to produce accurate results compared with floating point (FP). However, it brings unacceptable long latency.

In addition, arithmetic units also influence the accuracy of MAC and GEMM.
On the one hand, for multiplier, work~\cite{kim2016dynamic} concluded that the inaccuracy is maximal when the multiplication result is zero using the XNOR gate. So it removes near-zero operands. However, it works only if we know the value of operands in advance.
Work~\cite{zhakatayev2018sign} proposed a new sign-magnitude data format to optimize multiplier, but its adder can only compute one input in one adder, which is unfriendly for computing pipeline and increases the latency. Therefore, we utilize the sign-magnitude data format to improve the accuracy of multiplier, and redesign the adder for this format.

On the other hand, for adder, many works~\cite{sim2017new, jenson2016deterministic} lacked the bipolar non-scaled addition for MAC and GEMM in DNNs, without saying for sign-magnitude format. Work~\cite{li2017neural, faraji2019energy} proposed a separated adder for the sign-magnitude format, but it is greatly influenced by overflow problem. Therefore, we design new adder to mitigate overflow problem.



\subsection{Challenges and Motivation}\label{sec:challenges}

\textbf{Challenge1: High Data Precision Requires Long Computing Latency.}
In SC, high data precision is often required to obtain high accuracy. Specifically, for deterministic SC \cite{qian2010architecture}, to process m n-bits inputs, $2^{m \times log_2n}$-bit bitstreams must be generated.
Obviously, the computing time increases exponentially, leading to unacceptable latency. 
To shorten latency, we propose a block-based architecture, namely BSC, which divides inputs into blocks and executes blocks in parallel. In this way, to obtain high computing accuracy, the accuracy of arithmetic units in every block must be guaranteed.


\textbf{Challenge 2: Correlation Problem in Basic XNOR Multiplier and OR-tree Adder.}
In SC, multiplier and adder are two critical arithmetic units. As the basic circuits, XNOR multiplier and OR-tree adder \cite{li2013computation} cannot obtain high accuracy with bipolar because of strong correlation among inputs \cite{lee2018correlation}. 
If inputs are all independent, those operations can get a high accuracy. However, there are often correlations among inputs. For example, the addition of two bipolar inputs $a + b = c$ can be computed correctly using OR gate if and only if their corresponding bitstreams A and B are exclusive. 

To mitigate the correlation issue, the work~\cite{zhakatayev2018sign} proposed `sign-magnitude' format, which expresses data using absolute value with an additional sign bit in front, to optimize XNOR multiplier. 
However, their implementation in a sequential manner still faces long latency due to the single-input adder. To shorten the latency, the parallelism might be introduced but we may need to redesign the arithmetic units to adapt to this format and can handle multiple inputs for addition.



\begin{figure}[t]
\center
\includegraphics[width=1.0\linewidth]{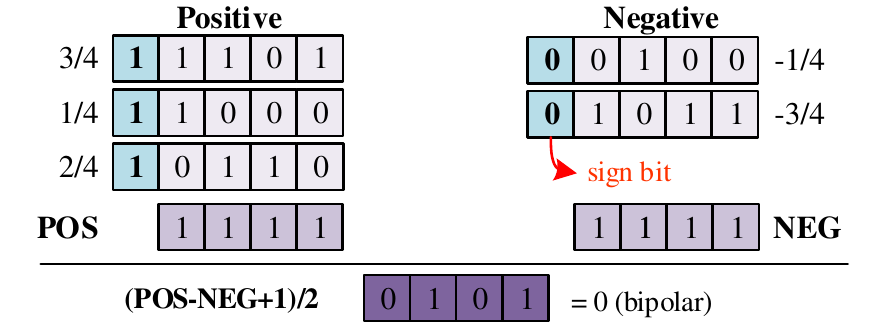}
\captionsetup{font={small}}
\caption{An example of multiple inputs addition with separated adder. There are total 5 inputs, 3 positive inputs and 2 negative inputs.}
\label{fig:example_partial_adder}
\end{figure}

\textbf{Challenge 3: Separated Adder Suffers from Severe Overflow Problem.}
For sign-magnitude data, we do multiplication using XNOR gate for sign bit, and AND gate for magnitude.
For addition, if all the data have the same sign, they can be added using an unipolar adder, and the sign of the result can be obtained directly. In MAC and GEMM operations, multi-input adders are needed.
However, the existing studies face the overflow issue severely to implement multi-input adders.
For example, in~\cite{li2017neural}, inputs are divided into positive and negative parts according to sign bits. It performs an OR tree for the summation of $POS$ and $NEG$. Then it utilizes scaled addition to compute $\frac{POS-NEG+1}{2}$, which exactly is the expression of bipolar $POS-NEG$.
However, the mean absolute error (MAE) of the separated adder is high (shown in Section \ref{sec:eva_adder}). 
Because the results of $POS$ and $NEG$ are often overflow for multiple inputs, which lead to $POS = 1$ and $NEG = 1$.
So $\frac{POS-NEG+1}{2}=1/2$ (i.e., 0 in bipolar). We give an example in Figure \ref{fig:example_partial_adder} to explain this phenomenon clearly.
So, parallel computation without input scaling will magnify the overflow problem, which leads to huge accuracy loss. 

Therefore, we propose an optimized merged adder to mitigate the correlation and overflow problems to improve the computation accuracy for multi-input adders inside blocks. 



\begin{figure*}[htbp]
\flushleft
\includegraphics[width=1\linewidth]{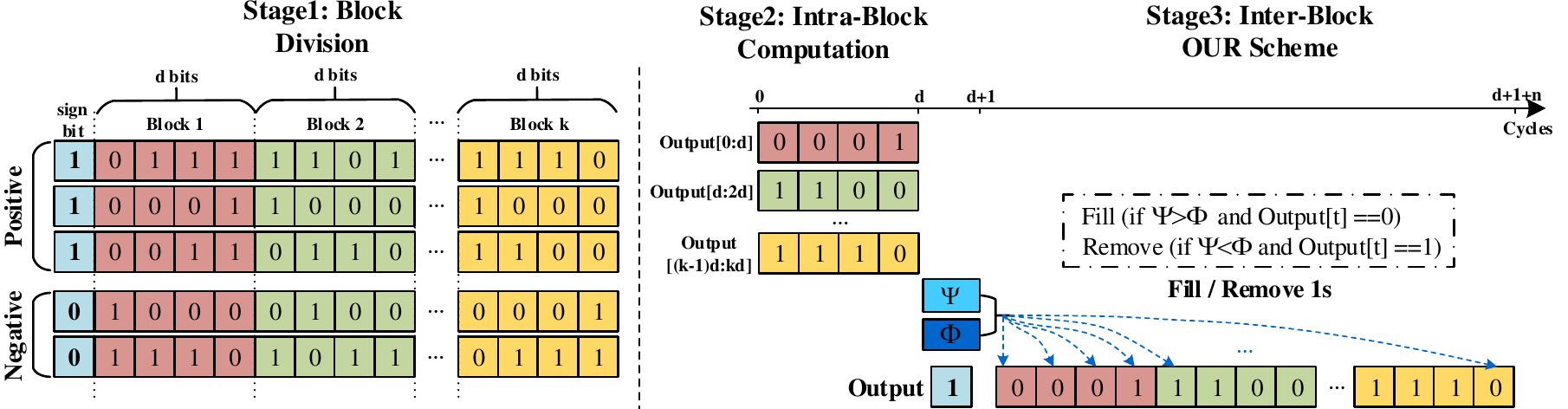}
\captionsetup{font={small}}
\caption{Overview of BSC. There are 3 stages: (1) block division; (2) intra-block computation in parallel; (3) revise output.}
\label{fig:overview}
\end{figure*}

\section{Design of the BSC architecture}
\subsection{Overview of BSC}
In order to improve accuracy and reduce computing latency, we propose a block-based SC architecture, namely BSC, which divides inputs into blocks and executes multiplication and addition in parallel. In this way, the computing latency is obviously reduced. 
Moreover, we utilize the sign-magnitude format to mitigate correlation problem for multiplier.
Meanwhile, we design accumulator-based adder to ensure intra-block addition accuracy (see Section \ref{sec:ACCADD}) and OUR scheme (see Section \ref{sec:revision_scheme}) to solve inter-block inaccuracy problem. Besides, a heuristic strategy of block division (see Section \ref{sec:bnum}) is proposed to determine the number of block. This strategy takes all accuracy, latency and energy into consideration. Figure \ref{fig:overview} illustrates the overview of our proposed BSC architecture. We introduce its details as follows.

\subsection{Intra-Block: Accumulator-based adder}\label{sec:ACCADD}
\begin{figure}[t]
\center
\includegraphics[width=0.98\linewidth]{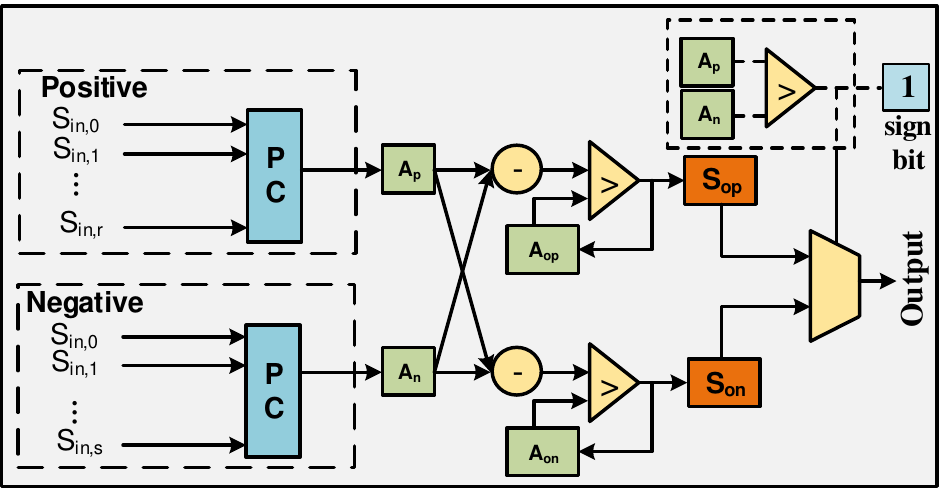}
\captionsetup{font={small}}
\caption{Circuit of intra-block accumulator-based adder.}
\label{fig:accummulated_adder}
\end{figure}

For multiplier, we directly apply XNOR+AND gates between inputs to mitigate correlation of bipolar computing. 
As indicated in Section~\ref{sec:challenges}, traditional adders suffer from correlation problem in the OR adder and overflow problem in the separated adder. In order to address these problems, inside every block, we design a new accumulator-based adder for sign-magnitude format as demonstrated in Figure \ref{fig:accummulated_adder}.
The new adder takes advantage of uNSADD adder \cite{wu2020ugemm} to compare the real accumulated `1's in inputs and output to determine the output bit. In every cycle, the accumulators of positive and negative parts are subtracted to calculate real accumulated `1's so far. This merged accumulator mitigates the impacts of correlation and quick overflow problems.
Meanwhile, the adder maintains the uniform of output, which decreases the effects of correlation for next computation.




Firstly, all inputs are entered into the parallel counter (PC). In every cycle, one bit of every input is accumulated into accumulators in parallel. The accumulation is processed separately for positive and negative parts to their own accumulators (marked as $A_p$ and $A_n$ in the figure). 
Secondly, subtract two accumulators. In this stage, we do not know the sign of output, so we use circuits to calculate $A_p - A_n$ and $A_n - A_p$, respectively. This step aims to get the number of accumulated `1's of all inputs so far.
Then, one bit of two temporal outputs $S_{op}$ and $S_{on}$ are obtained by comparison.
Finally, after $n$ cycles, we compute the sign bit and select the correct output result from $S_{op}$ and $S_{on}$ according to $A_p$ and $A_n$.
The temporal $S_{op}$ and $S_{on}$ are computed based on the comparison of $A_p - A_n / A_n - A_p$ and $A_{op}/A_{on}$. 
If $A_p - A_n$ is larger than $A_{op}$, which accumulates the number of `1's in the temporal output so far, then $S_{op}[t] = 1$. Otherwise, $S_{op}[t] = 0$, where t is the index of the temporal output. Then, $A_{op} = A_{op} + S_{op}[t]$. Initially, $A_{op} = 0$. The same is true for $S_{on}$.

\begin{figure}[b]
\flushleft
\includegraphics[width=1.0\linewidth]{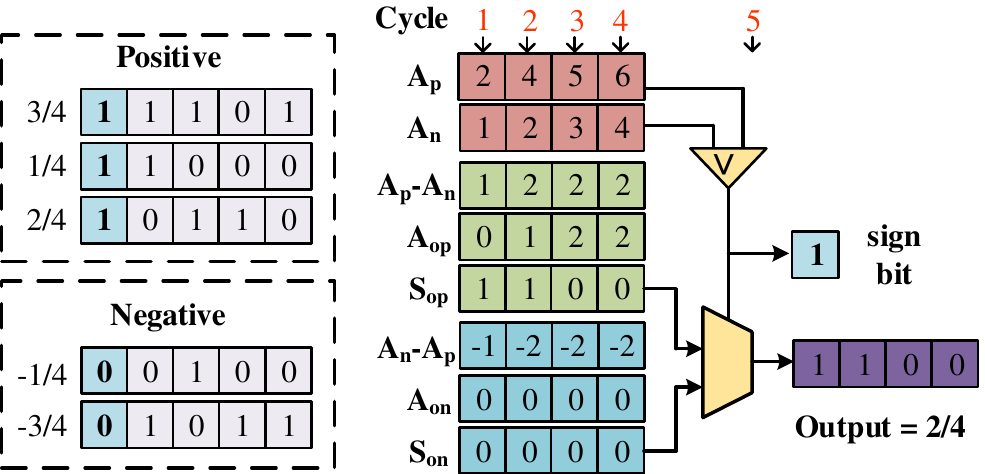}
\captionsetup{font={small}}
\caption{An example of accumulator-based adder with 5 inputs.}
\label{fig:exp_accumulator_adder}
\end{figure}

Now, we provide an example (shown in Figure \ref{fig:exp_accumulator_adder}) to explain the accumulator-based adder. In example, there are five inputs. $A_p$ and $A_n$ are accumulated based on the inputs in every cycle.
For example, in Cycle 2, $A_p = 4$ means the first two bits of positive inputs have total four bits.
Then $A_p - A_n$ and $A_n - A_p$ are calculated respectively. $A_{op}$ and $A_{on}$ are set as 0 initially, and accumulate `1's according to $S_{op}$ and $S_{on}$ in every cycle.
Compared with $A_{op}$ and $A_{on}$, $S_{op}$ and $S_{on}$ are determined.
Finally, we compare $A_p$ and $A_n$ in Cycle 5 to obtain sign bit 1 and select the corresponding output 1100.
We find that in this case the error is eliminated by the adder compared with the separated adder. However, it will bring pipeline stalls because we can't know the sign bit immediately. The pipeline stalls may be reduced with the number of block increasing.

\subsection{Inter-Blocks: Output Revision (OUR) Scheme}\label{sec:revision_scheme}
Although the intra-block adder can alleviate correlation and overflow problems well, the block division brings new inter-block inaccuracy problem. For multiplier, it does not have this problem since its inputs are XNORed and ANDed. However, for adder, the number of `1's in the output of each block may be more or less than the accurate result, which makes the errors among blocks are exacerbated. We give following two examples to explain the problem. 
%

\begin{figure}[t]
\flushleft
\includegraphics[width=0.98\linewidth]{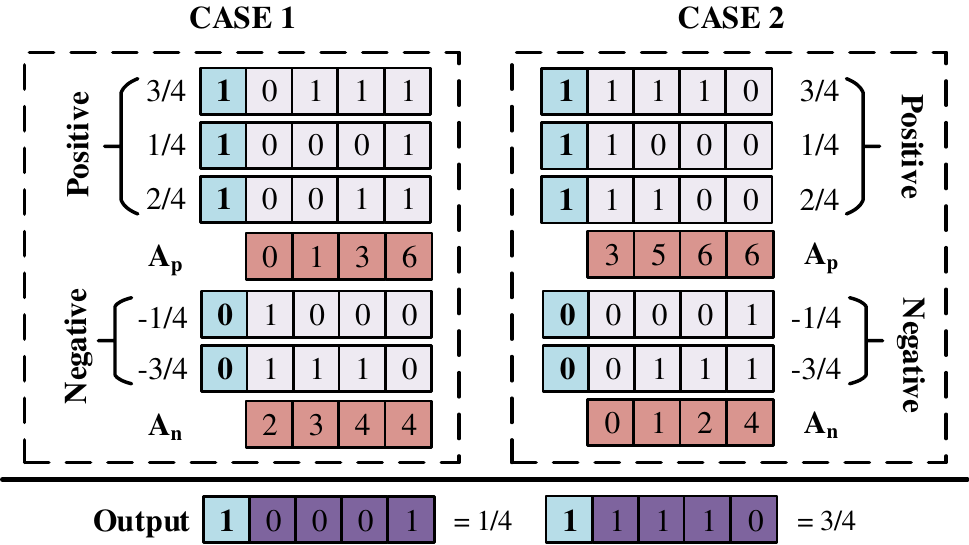}
\captionsetup{font={small}}
\caption{Error cases for accumulator-based adder.}
\label{fig:error_cases_accumulator_adder}
\end{figure}

In Figure \ref{fig:error_cases_accumulator_adder}, we list two cases of block computing with errors. The accurate results of two cases are both 2/4.
We omit detailed processes for more concise understanding. In CASE 1, most of `1's in the greater part (i.e., positive part in the figure) appear at the end of bitstreams, and most of `1's in the lower part (i.e., negative part in the figure) appear in the front of bitstreams. This case causes that the remained accumulated `1's finally have no place to fill them, which induces the number of `1's in this block is less than the accurate result. In CASE 2, the situation is exactly opposite. Most of `1's in positive part appear in the front of bitstreams and most of `1's in negative part appear at the end of bitstreams. In this case, there are more `1' in this block than that of the accurate result. Without the block division, this problem occurs only once in the whole bitstreams. However, because of block division, these cases may occur in every block, which induces inaccuracy among blocks. 

To solve this inter-block inaccuracy problem, we propose an OUR scheme to fill/remove `1's after parallel intra-block computation without any additional latency cost. OUR scheme is shown in the Stage3 of Figure~\ref{fig:overview}. 
First, we calculate the summation of $A_p$, $A_n$ and $A_o$ from all blocks. $A_p$, $A_n$ and $A_o$ identify the number of `1's of positive data, negative data, and output in every block. $\Psi$ and $\Phi$ in Figure \ref{fig:overview} refer to $|\sum A_p - \sum A_n|$ and $\sum A_o$. $\Psi$, $\Phi$ and temporal output, which consists of the output of every block, are compared and judged to execute `1's filling/removing process. If $\Psi$ is larger than $\Phi$, that means the accurate number of `1's is larger than the temporal output, so the filling process is launched. On the contrary, the removing process is executed.

\begin{figure}[t]
\center
\includegraphics[width=1.0\linewidth]{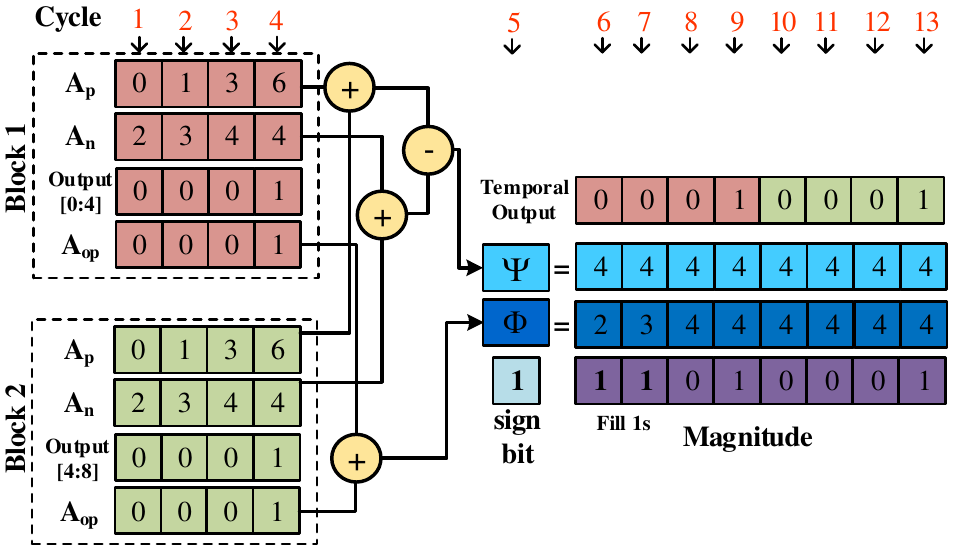}
\captionsetup{font={small}}
\caption{An example of OUR scheme for filling `1's. 
Inputs of two block are both from CASE 1 in Figure \ref{fig:error_cases_accumulator_adder}.
}
\label{fig:fill_loop_adder}
\end{figure}

Figure \ref{fig:fill_loop_adder} is an example to demonstrate the filling process of OUR scheme, in which the input bit-length is $n = 8$ and the number of block is $k = 2$. The inputs of both two blocks are from CASE 1 in Figure \ref{fig:error_cases_accumulator_adder}.
In Figure \ref{fig:fill_loop_adder}, intra-block computation is executed in Cycle 1-4, and filling process is performed from Cycle 5-13. 
$\Psi = 6+6-(4+4) = 4$, $\Phi = 1+1 = 2$ and the sign bit are computed in Cycle 5. In next two cycles, $\Psi$ is larger than $\Phi$ and temporal output[t] = 0, so we fill `1's at these two positions.
After Cycle 6 and 7, $\Psi$ is equal to $\Phi$, so the remaining bits can be output directly in every cycle.
Without OUR, the computation based on blocks also needs 13 cycles until every bit is output, while OUR makes the addition deterministic. 
As the number of blocks increases, the total cycles of intra-block computation can be further decreased.





\subsection{Heuristic Strategy for Block Division}\label{sec:bnum}
Regarding the proposed BSC, how to determine parallelism is a critical question. Because we divide bitstreams into blocks, the parallelism refers to the number of blocks. Aiming to select a good number of blocks, we propose a novel heuristic strategy to balance among accuracy, latency, and power.

\ding{172} On the accuracy side, the number of blocks impacts the accuracy of proposed intra-block adder. For our accumulator-based adder, the sign bit in every block is judged locally. If the local sign bit is inconsistent with the global sign bit, the output of this block may produce large errors (because it selects the opposite-sign result as output), which leads to even more overhead for the OUR scheme to fill/remove `1's. Moreover, more `1's filling/removing may break the uniform distribution of outputs when `1's/`0's become dense, which will exacerbate the correlation problem in later multiplication.
To ensure the local sign bit has a high probability of being consistent with the global sign bit, we propose a heuristic strategy.

For every block, we utilize the average value to represent the size of positive and negative parts. In this way, the judgment of two parts can be simplified to the judgement of two average bitstreams.
Next, we search for the minimal bit-length $d$ of blocks to make the correct probability of local sign bit judgment larger than $\theta$, which is an acceptable probability threshold.
The more `1's the bitstream has, the larger value this bitstream represents.
If the average bitstreams of positive and negative parts are represented as $AVG\_POS$ and $AVG\_NEG$, and the corresponding values are $p$ and $q$, the probability of $p \ge q$ can be expressed as: 

\begin{equation}
    P(p \ge q) = \sum_{i=0}^{d}[C_{d}^i p^i (1-p)^{d-i} \sum_{j=0}^{i} C_{d}^j q^j (1-q)^{d-j}], 
\end{equation}
where i refers to the number of `1's in $AVG\_POS$ and j refers to the number of `1's in $AVG\_NEG$. From the equation, i is always greater or equal to j.
Therefore, the correct probability of two average values is:

\begin{equation}
Probability=\left\{
\renewcommand\arraystretch{1.5}
\begin{array}{ll}
P(p \ge q) & {p \ge q}\\
1 - P(p \ge q)  & {p < q}\\
\end{array} \right.
\label{equa:prob}
\end{equation}
We enumerate all $p$ and $q$ range from 0 to 1 with step 0.1, and calculate the probability that the two average values can be correctly judged. Then, we can determine the minimal bit-length for every block where the correct probability is greater than $\theta$. Results will be shown in Section \ref{sec:explor_bnum}.

\ding{173} As we know, when the bitstreams are executed in parallel, the latency can be shorten multiple times according to the number of blocks. However, the hardware implementations become more complex. In order to trade-off between latency and hardware cost, a constraint is set. That is, the hardware power consumption of BSC must be lower than FP. 

Therefore, combining above two factors, the number of block can be determined heuristically, which brings better accuracy, lower latency and saved power. 

\section{Experiments}

\subsection{MAE Evaluation of Adders}\label{sec:eva_adder}
Adders as the dominant arithmetic circuits are first evaluated in this subsection. Note that all FP inputs are randomly generated and converted to bitstreams using Sobol generator. 

\begin{figure}[t]
\center
\includegraphics[width=0.97\linewidth]{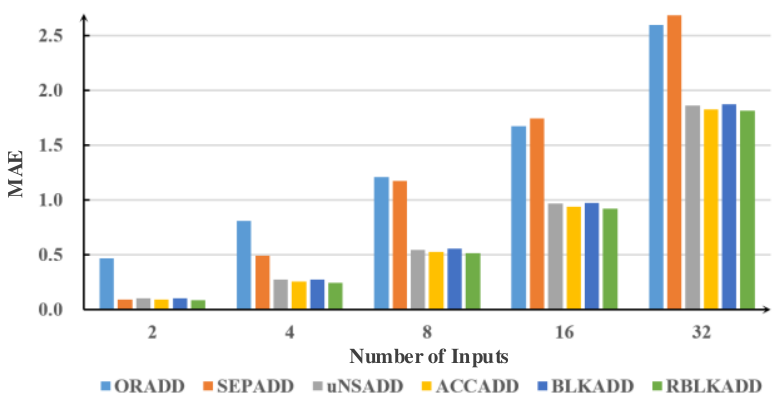}
\captionsetup{font={small}}
\caption{MAE results of adders with different number of inputs.}
\label{fig:exp_adder}
\end{figure}

\textbf{Setup.}
We demonstrate the comparison of six types of adders, including OR-tree adder \cite{li2013computation} (ORADD), separated adder \cite{li2017neural} (SEPADD), adder in uGEMM \cite{wu2020ugemm} for bipolar (uNSADD), our proposed intra-block adder (ACCADD), block-based adders without OUR (BLKADD), and with OUR (RBLKADD).  
Experiments show the MAE of adders under different number of inputs. Every bitstream is represented using 64 bits, and the number of block for BLKADD and RBLKADD is set as 4. 
Results are shown in Figure \ref{fig:exp_adder}.

\textbf{Results.}
Figure \ref{fig:exp_adder} demonstrates that with the number of inputs increasing, MAE shows an upward trend among adders because of the overflow problem, which is inevitable when the number of inputs get large. However, we find that SEPADD can't afford this overflow problem when the number of inputs is only 4, while its MAE is so low when executes 2-inputs addition. Even the MAE become higher than ORADD when the number of inputs is 16. For ORADD, correlation problem always keeps strong. Other 4 adders are much better than ORADD and SEPADD. Our proposed RBLKADD is the best because it alleviates the correlation and overflow problems using sign-magnitude format and intra-block ACCADD. So it is better than uNSADD. Moreover, it utilizes OUR scheme to make up errors caused by ACCADD and inter-block inaccuracy problem, which brings errors in BLKADD. 

\subsection{MAE Evaluation of GEMM}
This subsection evaluates the efficiency of proposed BSC architecture using GEMM operation.

\begin{figure}[b]
\center
\includegraphics[width=0.97\linewidth]{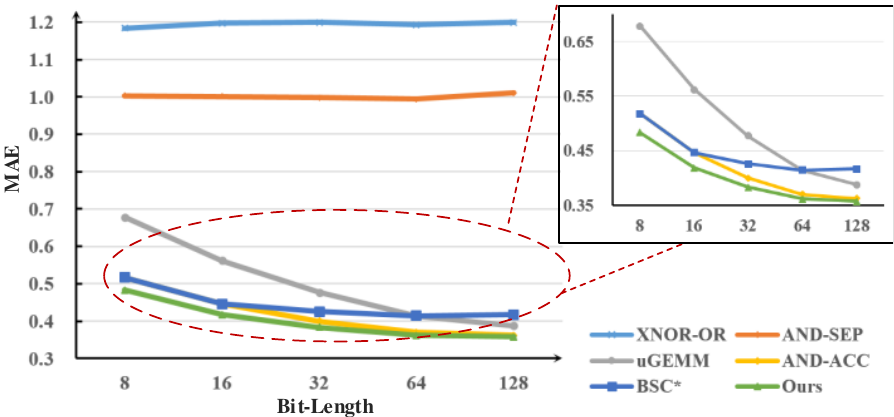}
\captionsetup{font={small}}
\caption{MAE results on GEMM with different input bit-length.}
\label{fig:exp_gemm}
\end{figure}

\textbf{Setup.} We compare XNOR-OR (XNOR multiplier and OR-tree adder for bipolar), AND-SEP (AND multiplier and separated adder for sign-magnitude), uGEMM (proposed by \cite{wu2020ugemm}), AND-ACC (AND multiplier and accumulator-based adder for sign-magnitude), BSC* (BSC without OUR) with our BSC. Experiments show GEMM results of two $16 \times 16$ matrices. We explore the impact of input bit-length on these methods. For BSC* and BSC, the bit-length $d$ in every block is set as 16. When the input bit-length is lower than $d$, the number of block is set as 1. Results are shown in Figure \ref{fig:exp_gemm}.

\textbf{Results.} For XNOR-OR and AND-SEP circuits, because their adders suffer from severe correlation and overflow problems when the number of input is 16 (see in Figure \ref{fig:exp_adder}), the MAE can't reduce with precision increasing. They always keep high MAE. For other 4 circuits, with the bit-length increasing, computation becomes more accurate. This is because these circuits can deal with multiple inputs well. 
As the same as Section \ref{sec:eva_adder}, our proposed BSC obtains best results because of proposed intra-block adder and inter-block OUR scheme.

\subsection{The Number of Block Exploration}\label{sec:explor_bnum}
\begin{figure}[b]
\center
\includegraphics[width=0.76\linewidth]{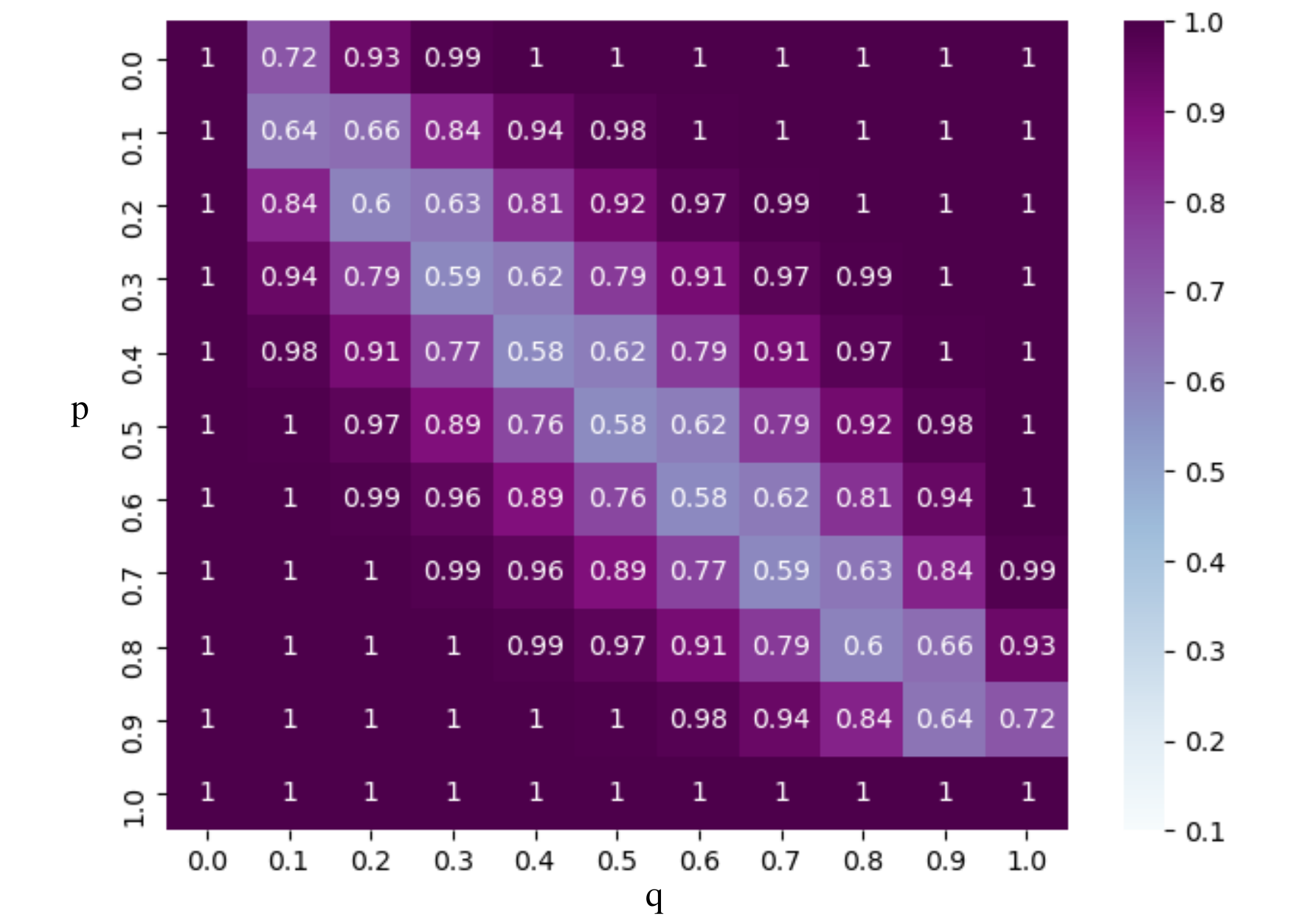}
\captionsetup{font={small}}
\caption{The correct probability distribution of size judgment when the bit-length of 2 bitstreams is 12.}
\label{fig:n12}
\end{figure}

In this section, the number of block is explored according to Section \ref{sec:bnum}. The number of block is determined heuristically from accuracy and latency-power aspects. 

a) From the accuracy aspect, we search the minimal bit-length of every block through Equation \ref{equa:prob}. We enumerate $p$ and $q$ in range [0, 1] with step 0.1, which includes total 11 $\times$ 11 = 121 combinations. 
Figure \ref{fig:n12} show the correct probability of size judgment using 12 bits. For example, when $p = 0.2$, $q = 0.3$, $P(0.2 < 0.3) = 0.63$.
The $d$ with an average probability greater than the threshold $\theta$ we set will be finally selected.
In our experiments, $\theta$ is set as 90\%. Through exhaustive search, the average probability becomes greater than $\theta$ when $d = 12$, its average probability is 90.28\%. Therefore, in our view, the bit-length of block should be larger than 12, that means there is more than 90\% probability that the size of two 12 bits bitstreams can be judged correctly.
This method can guarantee the addition accuracy inside every block, reduce the cost of OUR scheme and keep output relatively uniform.

b) From the latency-power aspect, a power constraint is set to find the adequate number of block. Table \ref{tab:explor_bnum} shows computing cycles, pipeline stalls, MAE and power consumption for MAC operation of two 16 dimensions vectors under different number of blocks. We only choose the number of block that can be divided by input bit-length exactly.
bitstreams are represented using 64 bits. SC circuits are implemented by Bluespec SystemVerilog, then compile to Verilog to evaluate power consumption on Vivado v2020.2.

\begin{table}[t]
\captionsetup{font={small}}
\caption{Results of the number of block exploration on MAC under BSC and BSC* circuits.}
\begin{minipage}{9cm}
\def\arraystretch{0.96}\tabcolsep 10pt
\def\thefootnote{a}\footnotesize
\begin{center}
\begin{tabular}{cc|c|c|c|c}
\hline
\multicolumn{2}{c|}{Design} & Cycles & Stalls & MAE & Power (W)  \\
\hline
\hline
\multirow{2}[1]{*}{1/64} & Ours & \multirow{2}[1]{*}{130} & \multirow{2}[1]{*}{64} & 0.314  & 2.17   \\
  & BSC* &   &   & 0.325  & 2.17  \\
\hline
\multirow{2}[1]{*}{2/32} & Ours & \multirow{2}[1]{*}{98} & \multirow{2}[1]{*}{32} & 0.362  & 3.35  \\
  & BSC* &   &   & 0.390  & 3.22   \\
\hline
\multirow{2}[2]{*}{4/16} & Ours & \multirow{2}[2]{*}{82} & \multirow{2}[2]{*}{16} & 0.305  & 5.74   \\
  & BSC* &   &   & 0.361  & 5.35   \\
\hline
\multirow{2}[2]{*}{8/8} & Ours & \multirow{2}[2]{*}{74} & \multirow{2}[2]{*}{8} & 0.346  & 10.86   \\
  & BSC* &   &   & 0.442  & 9.96   \\
\hline
\multirow{2}[2]{*}{16/4} & Ours & \multirow{2}[2]{*}{70} & \multirow{2}[2]{*}{4} & 0.368  & 21.02   \\
  & BSC* &   &   & 0.519  & 19.09   \\
\hline
\multirow{2}[2]{*}{32/2} & Ours & \multirow{2}[2]{*}{68} & \multirow{2}[2]{*}{2} & 0.332  &   41.31\\
  & BSC* &   &   & 0.533  &   37.31\\
\hline
\multirow{2}[2]{*}{64/1} & Ours & \multirow{2}[2]{*}{67} & \multirow{2}[2]{*}{1} & 0.352  &   81.02\\
  & BSC* &   &   & 0.562  &   72.90\\
\hline
\end{tabular}%
\end{center}
\end{minipage}
\label{tab:explor_bnum}%
\end{table}

The first column of Table \ref{tab:explor_bnum} identifies different number/bit-length of blocks and methods. 
Results show that compared with BSC*, our BSC can improve the accuracy without any additional cycles. 
But the power grows greatly with the parallelism increasing. For FP computing, the power of MAC for two 16 dimensions vectors is 34.6W. So we select the number of block lower than 32 for 64-bits inputs. 

Therefore, combine the results of a) and b), we choose $k = 4$ and $d = 16$ as the recommended number and bit-length of block when data is represented as 64 bits.
In this way, the computing latency can be reduced as much as possible. And we can achieve over 6$\times$ power saving than FP circuit.
For other data precision, the process is similar.

\subsection{Evaluation of Latency}
In this subsection, we compare the cycles, pipeline stalls and MAE of 6 circuits for MAC. 
Table \ref{tab:exp_lat} demonstrates the results of two 16 dimensions vectors with 64-bits. $k$ is set as 4 in method BSC* and our BSC.

\begin{table}[t]
\captionsetup{font={small}}
\caption{Evaluation results of latency on MAC.}
\begin{minipage}{9cm}
\def\arraystretch{1.15}\tabcolsep 17pt
\def\thefootnote{a}\footnotesize
\begin{center}
\begin{tabular}{c|c|c|c}
\hline
Design & MAE & Cycles & Stalls \\
\hline
\hline
XNOR-OR & 1.097  & 65 & 0 \\
\hline
AND-SEP & 0.937  & 65 & 0 \\
\hline
uGEMM & 0.362  & 65 & 0\\
\hline
AND-ACC & 0.315  & 130 & 64 \\
\hline
BSC* & 0.361  & 82 & 16 \\
\hline
Ours & 0.305  & 82 & 16 \\
\hline
\end{tabular}%
\end{center}
\end{minipage}
\label{tab:exp_lat}%
\end{table}

From Table \ref{tab:exp_lat}, we see that our BSC will produce a little higher cycles and pipeline stalls than first 3 circuits. But we achieve 3.6$\times$, 3.1$\times$ and 1.2$\times$ accuracy improvement than them. 
The improvement can be larger in MLP implementation because errors will be accumulated among thousands of MAC. We also observe that BSC has similar MAE than AND-ACC, but BSC saves lots of cycles because of parallel computing. Therefore, in the next case study of MLP, we remove comparison with this method.
Compared with BSC*, ours can improve accuracy without additional cycles.
Besides, BSC implements deterministic adder because of OUR, its latency is much lower than deterministic circuits with $2^{16*log_264} = 2^{96}$ cycles.

\begin{table}[t]
\captionsetup{font={small}}
\caption{The accuracy of 3-layers perceptron with 32 and 64 hidden neurons under FP circuit and different SC circuits.}
\begin{minipage}{9cm}
\def\arraystretch{1.3}\tabcolsep 4.5pt
\def\thefootnote{a}\footnotesize
\begin{center}
\begin{tabular}{c|cccccc}
\hline
Design & FP & XNOR-OR & AND-SEP & uGEMM & BSC* & Ours \\
\hline
\hline
Accuracy & 96.1\% & 10.0\% & 21.7\% & 85.1\%  & 93.4\% & 95.4\% \\
\hline
Acc.loss & - & 86.1\% & 74.4\% & 11.0\%  & 2.7\% & 0.7\% \\
\hline
\end{tabular}%
\end{center}
\end{minipage}
\label{tab:mlp}%
\end{table}

\subsection{MLP Implementation}
Finally, we implement MLP as a study case to show the strength of our proposed BSC on ML.

We develop a custom SC simulator integrated with PyTorch to implement these circuits. The MLP is 3-layers perceptron with 32 and 64 hidden neurons. We first train the MLP model 30 epochs using FP, then implement MLP using above 5 circuits. The inference accuracy (shown in Table \ref{tab:mlp}) is evaluated on the MNIST dataset.
Results show that our method can achieve the best inference accuracy. And there is only 0.7\% accuracy gap compared with FP.


\section{Conclusion}
This work present a block-based SC architecture, namely BSC, aiming at improving the accuracy of SC arithmetic circuits, reducing computing latency and saving energy. BSC divides inputs into blocks, then they are executed in parallel. We propose a novel intra-block accumulator-based adder and inter-block output revision (OUR) scheme to improve accuracy. Moreover, we propose a heuristic strategy to determine the number of block, which takes accuracy, latency and power consumption into consideration. Results show that our method achieves over 10\% higher accuracy than existing methods, and saves over 6$\times$ power consumption.

\section*{Acknowledgement}
We gratefully acknowledge the support of National Nature Science Foundation of China (NSFC) 61972154 and Shanghai Science and Technology Commission Project 20511101600 to Yuhong Song, Edwin Hsing-Mean Sha, Qingfeng Zhuge, Rui Xu, Yongzhuo Zhang.

\bibliographystyle{ieeetr}
{\scriptsize
\bibliography{ref}}

\end{document}